\documentclass{article}

\usepackage{microtype}
\usepackage{graphicx}
\usepackage{subfigure}
\usepackage{booktabs} 

\usepackage{hyperref}

\usepackage[accepted]{icml2023}

\usepackage{amsmath}
\usepackage{amssymb}
\usepackage{mathtools}
\usepackage{amsthm}

\usepackage[capitalize,noabbrev]{cleveref}

\theoremstyle{plain}

\theoremstyle{definition}

\theoremstyle{remark}

\usepackage[textsize=tiny]{todonotes}

\icmltitlerunning{Surround-View Vision-based 3D Detection for Autonomous Driving: A Survey}

\begin{document}

\twocolumn[
\icmltitle{Surround-View Vision-based 3D Detection for Autonomous Driving: A Survey}



\icmlsetsymbol{equal}{*}

\begin{icmlauthorlist}
\icmlauthor{Apoorv Singh}{Motional}
\icmlauthor{Varun Bankiti}{Motional}
\end{icmlauthorlist}

\icmlaffiliation{Motional}{Motional, Pittsburgh, USA}

\icmlcorrespondingauthor{Apoorv Singh}{apoorv.singh@motional.com}
\icmlcorrespondingauthor{Varun Bankiti}{varun.bankiti@motional.com}

\icmlkeywords{Machine Learning, Computer Vision, Autonomous Driving, Object Detection, ICML}

\vskip 0.3in
]



\printAffiliationsAndNotice{} 

\begin{abstract}
Vision-based 3D Detection task is fundamental task for the perception of an autonomous driving system, which has peaked interest amongst many researchers and autonomous driving engineers. However, achieving a rather good \emph{3D} BEV (Bird's Eye View) performance is not an easy task using \emph{2D} sensor input-data of monocular cameras. In this paper we provide a literature survey of the existing Vision-Based 3D detection methods, focused on autonomous driving. We have made detailed analysis of over $60$ papers leveraging Vision BEV detection approaches and binned them in different sub-groups for easier understanding of the common trends. Moreover, we have highlighted how the literature and industry trend have moved towards surround-view image based methods and note down thoughts on what special cases these surround-view methods address. In the conclusion, we provoke thoughts of 3D Vision techniques for future research based on the shortcomings of the current methods including direction of collaborative perception. Regularly updated summary can be found at \emph{https://github.com/ApoorvRoboticist/VisionBEV\\DetectionSurvey}.
\end{abstract}

\section{Introduction}

\begin{figure}[ht]
\vskip 0.2in
\begin{center}
\centerline{\includegraphics[width=\columnwidth]{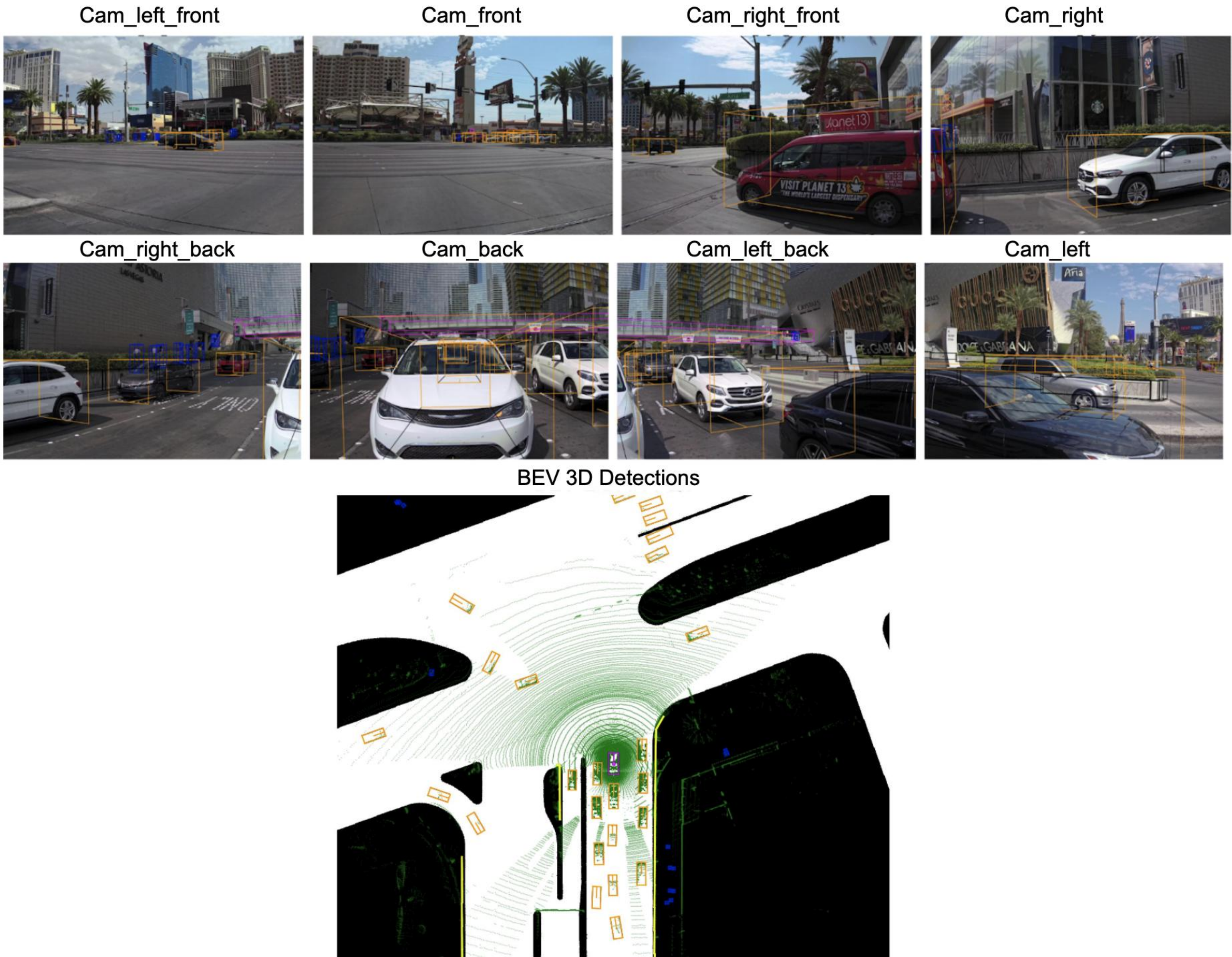}}
\caption{Surround-view Image 3D Detector in autonomous driving. Ground-truth 3D boxes overlaid over surround-images in perspective view (top); Ground-truth 3D boxes overlaid over BEV HD Map (bottom), with ego car in pink.}
\label{bev-output}
\end{center}
\vskip -0.2in
\end{figure}

\label{structure}
\begin{figure}[ht]
\vskip 0.2in
\begin{center}
\centerline{\includegraphics[width=\columnwidth]{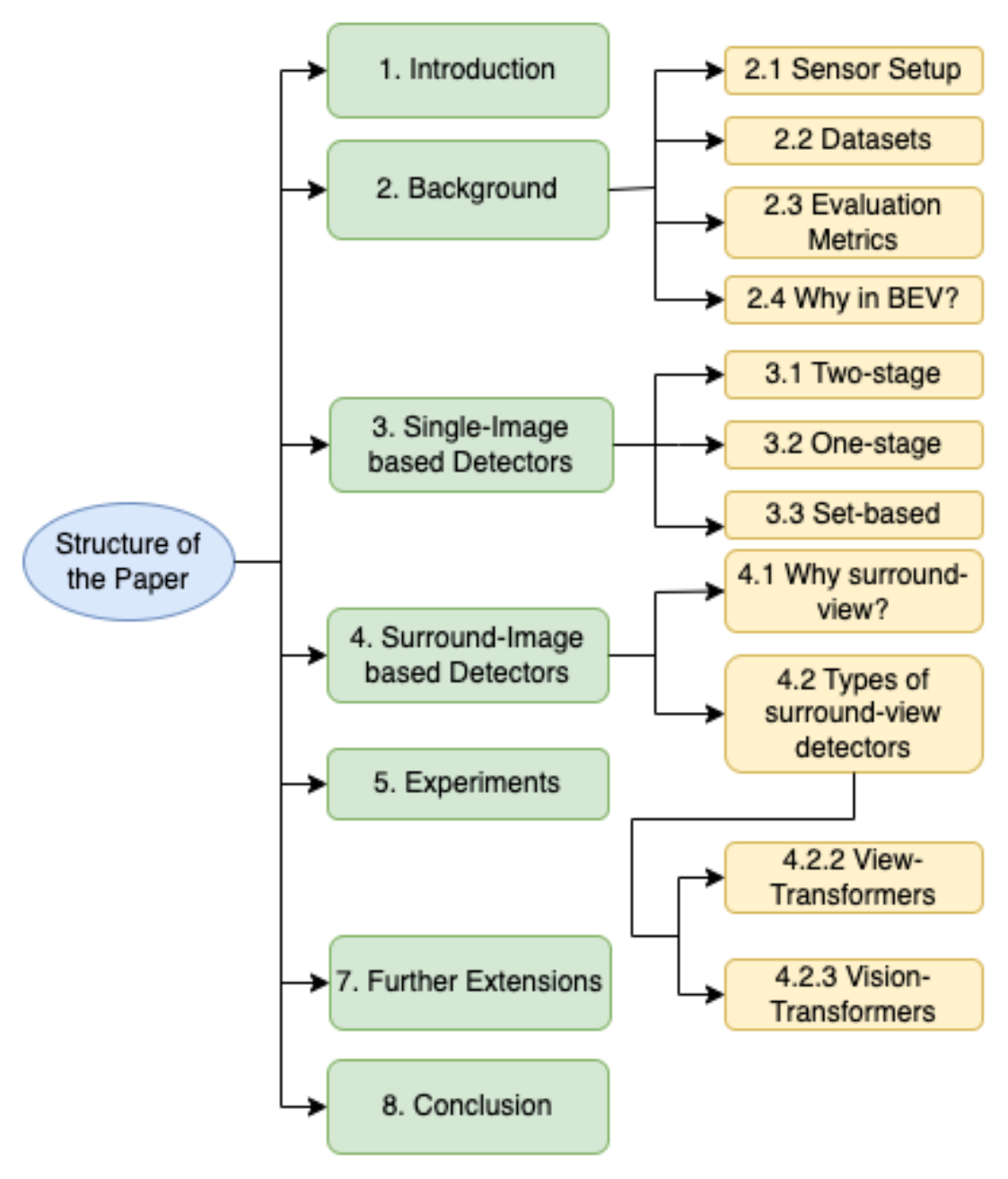}}
\caption{Structure of this Survey Paper.}
\label{paper-structure}
\end{center}
\vskip -0.2in
\end{figure}

Object detection is a trivial task for humans. Pretty much any teenager can look at the scene out of the car's windscreen and place all the agents, dynamic or static, in a mental BEV (Bird's Eye View) map. This virtual map may include per-agent information as, but not limited to, center coordinates, dimensions, orientation angle etc. However, teaching this to a computer has been a nearly impossible task until the turn of the last decade. This task entails identifying and localizing all the instances of an object (like cars, humans, street signs, etc.) within the field of view as shown in \cref{bev-output}. Similarly classification, segmentation, dense-depth estimation, motion prediction, scene understanding etc., are some of the other fundamental problems in the field of computer vision. \\ \\
Early object detection models were build on hand-crafted feature extractors such as Viola-Jones detector \cite{viola01}, Histogram of Oriented Gradients (HOG) \cite{dalal05} etc. These were SoTA (State-of-the-art) of their time, however, compared to the current methods these methods are slow, inaccurate and not scalable on generic datasets. Introduction of convolutional neural network (CNNs) and deep learning for image classification changed the landscape of visual perception. CNN's use in ImageNet Large Scale Visual Recognition Challenge (ILSVRC) 2012 challenge by AlexNet \cite{Krizhevsky12} has inspired further research on CNNs in the computer vision industry. Mainstream applications for 3D object detection lies around autonomous driving, mobile-robotic vision, security cameras etc. Limited Field-of-view (FOV) of cameras has led researchers to the next breakthrough problem-statement of \emph{how to leverage views from multiple cameras to reason the $360^\circ$ surroundings.} \\ \\
This survey on Surround-view Vision-based 3D object detection provides a comprehensive review of deep learning based methods and architectures in the recent past. The main contributions of this paper are as follows:
\begin{itemize}
\item This paper provides an in-depth analysis of major single-view detector baselines that inspired surround-view  detector research in 3D object detection task using cameras.
\item This paper provides further analysis of major surround-view detector trends currently in-development in the computer-vision community; thereby categorizing them for readers to follow-through easily. 
\item This paper provided detailed background on evaluation metrics and datasets used to evaluate and compare above methods.
\item This paper makes a detailed analysis about the remaining problems and introduce several potential research directions about the BEV 3D image object detectors, hence opening a possible door for future research.
\end{itemize}

Rest of the paper is organized as follows: We first look at the background information required to understand autonomous driving 3D detections viz., evaluation metrics, datasets, annotations etc. in \cref{background_section}. Then, we introduce single-image based detection methods and SoTA approaches that inspired surround-view detectors approaches in \cref{single_image_detectors_section}. In \cref{surround_image_detector_section}, we dive into details for surround-view based detections approaches focused on autonomous driving. We then report and analyse performance of these approaches on our previously defined metrics in \cref{experiments_key}. Then in \cref{further_extension_section}, we report possible research extensions on surround-view object detection methods that may enlighten future research. Finally in \cref{conclusion_section}, we conclude the paper.

\section{Background}
\label{background_section}
In order to cover the basics required to understand 3D BEV object detection tasks, we discuss four aspects: Sensor setup on an autonomous vehicle (AV); frequently used datasets; common evaluation metrics for detection task in autonomous driving, and Why Bird's Eye View (BEV) is important for an AV camera perception? 
\subsection{Sensor Setup}
\begin{figure}[ht]
\vskip 0.2in
\begin{center}
\centerline{\includegraphics[width=\columnwidth]{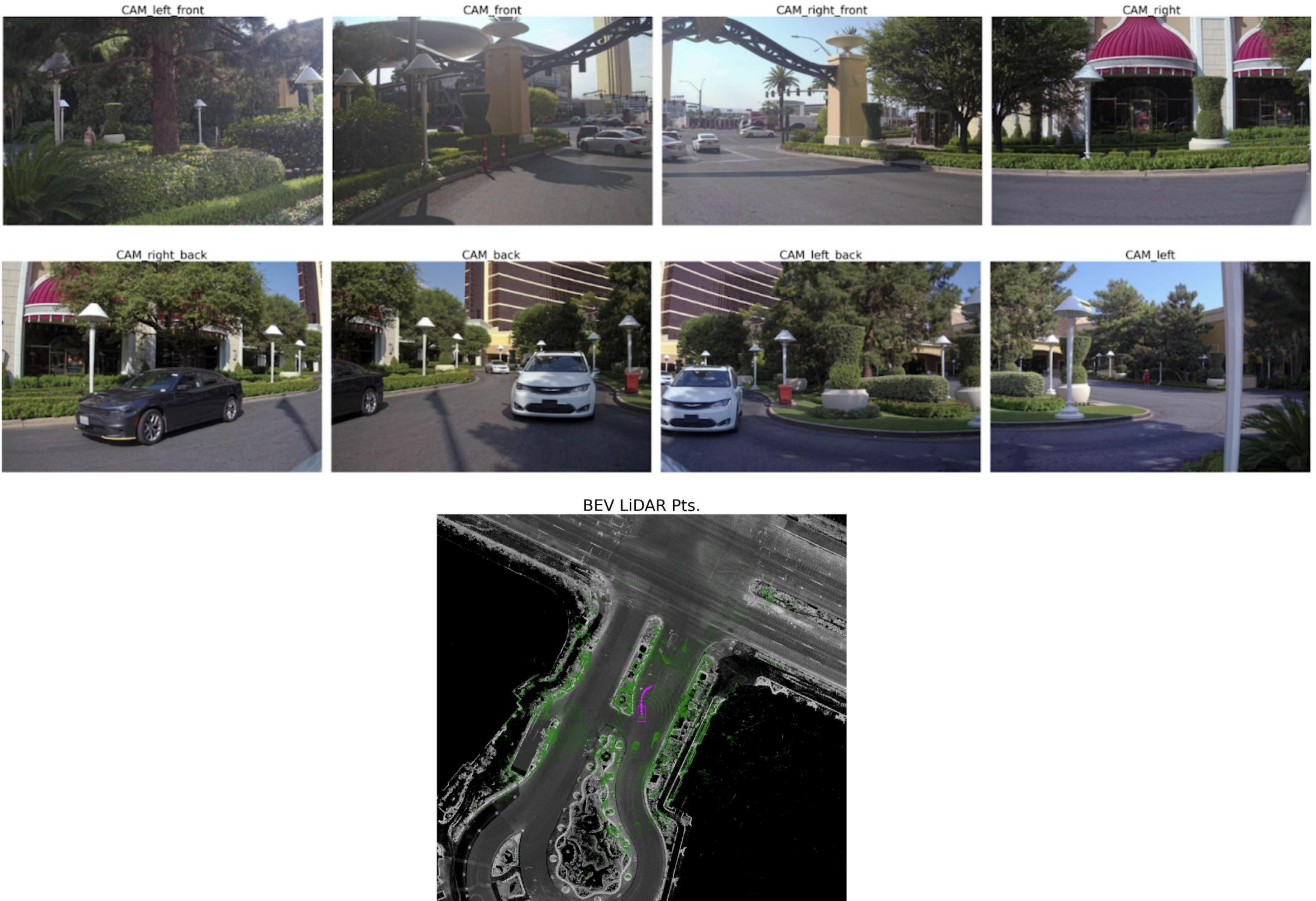}}
\caption{Surround-view 8 camera images (top); LiDAR Point Cloud overlayed over an HD Map (bottom). Key: Green points: LiDAR point cloud; Pink box: Ego Autonomous vehicle; Gray map: Pre-computed HD map with color intensity.}
\label{sensor-data}
\end{center}
\vskip -0.2in
\end{figure}

Before we even look at how cameras are setup in an autonomous vehicle (AV), lets try to understand why we need cameras at the first place. 
Cameras have the most densely packed information compared to other sensors, making them one of the most challenging, sensors to extract information from in an AV, however the most useful at the same time. To understand this mathematically let us first look at the number of data points in each of the visualizations as shown in \cref{sensor-data}. Take these data points (floating point numbers) as the input to the perception algorithm for a sensor to cover $360^{\circ}$ view, that is responsible to make decisions for an AV.\\
\label{submission}
\textbf{Let's start with a multi-camera:}\\  
Number of cameras: $8$; \\
Number of pixels per camera: $~2,000*3,000$ (image pixel resolution: width*height);\\ Representation of a pixel: 3 (three channeled RGB value).\\ \emph{This brings total parameters to: $8*2000*3000*3 = ~144M$ float numbers!} \\
\textbf{Similar comparison with a LiDAR now:}\\
Number of LiDAR points in a point cloud: ~250,000;\\
Representation of each LiDAR point: 4 (3D coordinate i.e. \emph{x, y, z} and reflectance).\\ 
\emph{This brings total parameters to: $250,000*4 = 1M$ float numbers!}

\begin{figure}[ht]
\vskip 0.2in
\begin{center}
\centerline{\includegraphics[width=\columnwidth]{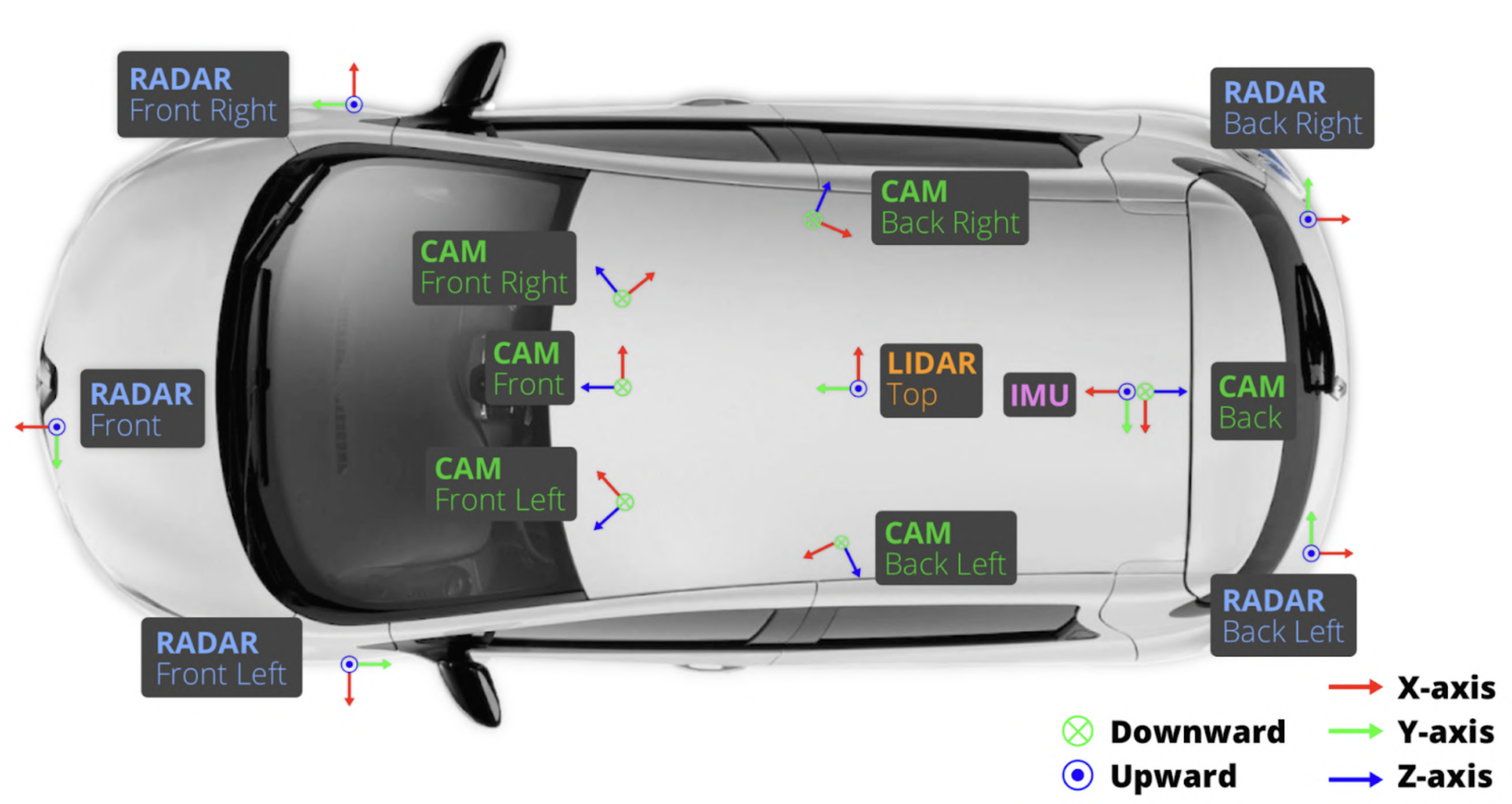}}
\caption{Sensor setup for an Autonomous vehicle in nuScenes \cite{nuscenes_20} benchmark dataset.}
\label{sensor_setup}
\end{center}
\vskip -0.2in
\end{figure}

These numbers and visualizations as in \cref{sensor-data} should be enough to prove our point of \emph{
“the key role cameras play in the AV perception to perceive the environment.”}\\ \\
A camera is one of the least expensive sensor too compared to other laser-based sensors. However, cameras are spectacularly better for detecting long-range objects and extracting vision-based road cues like state of traffic lights, stop signs etc, compared to any other laser-based sensor.
Setup of surround-view cameras on an AV may vary depending on different autonomous car manufacturer, but typically there are $6\sim12$ cameras per vehicle. These many cameras are needed to cover the entire surrounding 3D scene. We are limited to use cameras with normal FOV (Field of view) otherwise we may get image distortions that are beyond recovery, like with Fish-eye cameras (Wide FOV), which are only good for up to few tens of meters. A perception sensor setup in one of the most cited benchmark-dataset, nuScenes \cite{nuscenes_20} in the AV space can be seen in \cref{sensor_setup}.

\subsection{Datasets}
nuScenes\cite{nuscenes_20}, KITTI\cite{kitti12}, Waymo Open Dataset (WOD)\cite{wod_20} are the three most commonly used datasets for 3D BEV object detection task. Apart from them, H3D\cite{h3d_19}, Lyft L5\cite{lyftl5_19} and Argoverse\cite{chang_2019} can also be used for BEV perception tasks.
nuSences contains 1000 scenes with a duration of 20 seconds each. They contain six calibrated images covering the $360^{\circ}$ view of the road. Sensor setup with nuScenes can be seen in \cref{sensor_setup}. KITTI was the seminal work on autonomous driving dataset. It consists of a smaller sample of data compared to the more recent ones. Waymo Open Dataset (WOD) is another large-scale autonomous driving dataset with 798 sequences training, 202 validation and 150 testing sequences respectively. Argoverse 2 also contains 1000 scenes with LiDARs, stereo-imagery and ring-camera imagery a.k.a surround-cameras imagery. 
Detailed information of these dataset is given as in \cref{dataset-benchmark}.


\begin{table*}[t]
\caption{Information on benchmark dataset commonly used for 3D BEV Object Detection using cameras in autonomous driving.}
\label{dataset-benchmark}
\vskip 0.15in
\begin{center}
\begin{small}
\begin{sc}
\begin{tabular}{lcccccccccr}
\toprule
Dataset & Cameras  & Scenes & Train & Test & Boxes & Classes & Temporal & LiDAR & RADAR\\
\midrule
nuScenes    & 6 & 1,000 & 28,130 & 6,008 & 1.4M & 10 & $\surd$ & $\surd$ & $\surd$ \\
KITTI (3D)    & - & - & 7,418 & 7,518 & 200K & 3 & $\surd$ & $\surd$ & $\times$ \\
WOD    & 5 & 1,150 & 122,200 & 40,077 & 12M & 4 & $\surd$ & $\surd$ & $\times$ \\
Argoverse    & 7 & 113 & 39,384 & 12,507 & 993K & 15 & $\surd$ & $\surd$ & $\times$ \\
Lyft L5    & 6 & 366 & 22,690 & 27,468 & 1.3M & 9 & $\surd$ & $\surd$ & $\times$ \\
H3D    & 3 & 160 & 8,873 & 13,678 & 1.1M & 8 & $\surd$ & $\surd$ & $\times$ \\
\bottomrule
\end{tabular}
\end{sc}
\end{small}
\end{center}
\vskip -0.1in
\end{table*}

\subsection{Evaluation Metrics}
3D object detectors use multiple criteria to measure performance of the detectors viz., precision and recall. However, mean Average Precision (mAP) is the most common evaluation metric. Intersection over Union (IoU) is the ratio of the area of overlap and area of the union between the predicted box and ground-truth box. An IoU threshold value (generally 0.5) is used to judge if a prediction box matches with any particular ground-truth box. If IoU is greater than the threshold, then that prediction is treated as a True Positive (TP) else it is a False Positive (FP). A ground-truth object which fails to detect with any prediction box, is treated as a False Negative (FN). Precision is the fraction of relevant instances among the retrieved instances; while recall is the fraction of relevant instances that were retrieved.
\begin{equation}
Precision=TP/(TP+FP)
\end{equation}
\begin{equation}
Recall=TP/(TP+FN)
\end{equation}
Based on the above equations, average precision is computed separately for each class. To compare performance between different detectors (mAP) is used. It is a weighted mean based on the number of ground-truths per class. Alternatively F1 score is the second most common detection metric, which is defined as weighted average of the precision and recall. Higher \emph{AP} detectors gives better performance when the model is to be deployed at varied confidence thresholds, however higher \emph{max-F1} score detector is used when the model is to be deployed at a known fixed optimal-confidence threshold score. 
\begin{equation}
F1=2*Precision*Recall/(Precision+Recall)
\end{equation} \\ \\
In addition, there are a few dataset specific metrics viz., KITTI introduces Average Orientation Similarity (AOS), which evaluates the quality of orientation estimation of boxes on the ground plane. mAP metric only considers 3D position of the objects, however, ignores the effects of both dimension and orientation. In relation to that, nuScenes introduces TP metrics viz., Average Translation Error (ATE), Average Scale Error (ASE) and Average Orientation Error (AOE). WOD introduces Average Precision weighted by Heading (APH) as its main metric. It takes heading/ orientation information into the account as well. Also, given depth confusion for 2D-sensors like camera, WOD introduces Longitudinal Error Tolerant 3D Average Precision(LET-3D-AP), which emphasizes more on lateral errors than longitudinal errors in predictions.

\subsection{Why BEV (Bird's Eye View)?}
There are number of reasons why using 3D agent's representation in the Bird's Eye View makes more practical sense for autonomous driving:
\begin{itemize}
\item It makes fusion with the other $360^{\circ}$ sensors i.e. LiDARs and RADARs more natural as these laser-based sensors operate in the BEV space natively. 
\item If we operate in BEV, we can model temporal consistency of the dynamic scene much better. Motion compensation i.e. translation and rotation modeling in BEV agents is much more trivial compared to the perspective-view (camera-view). For example; In BEV view: Pose change depends just on the motion of the agent, whereas in perspective-view, pose change depends on the depth as well as the motion of the agent.
\item Scale of the objects are consistent in BEV space, but not so much in the perspective view. In perspective view objects appear bigger when they are closer to the view-point. Hence, BEV space makes it easier to learn range-agnostic scale features.
\item In an autonomous driving, downstream tasks after perception, like motion prediction and motion planning operate on the BEV space natively. It makes natural sense for all the software stacks to work in a common coordinate-view system on a robotic platform. 
\item Newly researched field, Collaborative perception which we will talk about in Section \cref{further_extension_section} also utilizes BEV representation for representing all the agents at a common coordinate system.
\end{itemize}

\section{Single-Image Based Detectors}
\label{single_image_detectors_section}
We have divided single-view image based object detection methods based on three categories: two-stage, single-stage and set-based detectors. However, we would like to mention pioneer works like Viola-Jones \cite{viola01}, HOG Detector \cite{dalal05}, Deformable Parts Model (DPM) \cite{dpm_14} which have revolutionized computer vision with PASCAL VOC challenge in 2009 \cite{pascal_07}. These approaches uses classical computer-vision techniques which relies on extracting human-designed heuristic features. \\
\subsection{Two-stage Detectors}
This is a class of detectors, divided into two stages. First stage is to predict arbitrary number of object proposals, and then in the second stage they predict boxes by classifying and localizing those object proposals. However, these proposals have inherent problem of slow inference time, lack of global context (even within the single image) and complex architectures. Pioneer work with two-stage approach are: Region-based fully convolution network (R-FCN) \cite{rfcn_16}, Feature Pyramid Network (FPN) \cite{fpn_16} and Mask R-CNN \cite{maskrcn_17} which are built upon R-CNN \cite{rcnn_13} line of work. There's also a parallel stream of work around Pseudo-LiDAR \cite{pseduolidar_18} in which dense-depth is predicted in the first stage, thereby converting pixels into a pseudo point-cloud and then LiDAR like detection head is applied for 3D object detection as done in Point-pillars \cite{pointpillar_18}. \\
\subsection{Single-stage Detectors}
YOLO \cite{yolo_15} and SSD \cite{ssd_15} opened the gate for single-stage detectors. These detectors classify and localize semantic objects in a single shot using dense-predictions. However, they rely heavily on post-processing Non-maximum Suppression (NMS) step to filter out duplicate predictions, as one of the over-head. Their dependence on anchor boxes heuristics was addressed in Fully Convolutional One-Stage Object Detection (FCOS) \cite{fcos_19} to predict 2D boxes based on center-pixel. Extension of this work is seen in FCOS3D \cite{fcos3d_21} where they address 3D object detection problem, by regressing 3D parameters per object. These methods still heavily rely on post-processing for duplicate detections with NMS. \\
\subsection{Set-based Detectors}
This approach removes hand-designed NMS using set-based global loss that forces unique predictions per-object via bipartite matching. The pioneer paper, DETR \cite{detr_20} started this chain of work. However it suffers through slow convergence which limited spatial resolution of the features. However, this issue was later addressed in Deformable-DETR \cite{deformable_20} method which replaces the original global dense atttention with deformable attention that only attends to a small set of sampled features to lower the complexity and thereby speeding up the convergence. Another approach to accelerate convergence is SAM-DETR \cite{sam_detr_22} which limits the search-space for attention module by using the most discriminative features for semantic-aligned matching. This line of work still has a CNN based backbone, however they use transformer \cite{transformer_17} based detection head.\\  \\
Above mentioned approaches operate per-camera, however autonomous driving application needs to address the entire $360^{\circ}$ scene which includes $6\sim12$ surround-cameras covering the entire spatial scene. Per-camera detections are generally aggregated using another set of NMS filtering to get rid of repeat detections originating from the camera overlap Field of View (FOV) regions. AVs need to maintain this long-range high FOV overlap to minimize the blind-spots in the short-range. Perspective view detections are lifted to BEV space either by regressing depth per objects or using heuristic based method, Inverse Perspective Mapping, by estimating ground-plane height.

\section{Surround-Image Based Detectors}
\label{surround_image_detector_section}
There are multiple applications of surround-camera based computer-vision (CV) systems like surveillance, sports, education, mobile phones, Autonomous Vehicles. Surround-view systems in sports are making a huge role in the sports analytics industry. It lets us record the right moment across the field at the right moment with the right viewing angle. Surround-view vision has also spread its application in class monitoring systems, which lets teachers give personalized attention to each student in the class. Nowadays it is hard to find any smart phone with a single camera. However, to limit the scope of this paper we will only focus on Autonomous driving based computer vision. \\ \\
A surround-view system makes use of features from different views to understand the holistic representation of the scene around the autonomous vehicle. A combination of any two or more cameras requires prior infrastructure work related to fixed sensor mountings and their calibration. Calibration of the camera simply means extracting the extrinsics transformation matrix between the two cameras. This camera matrix enables us to make one-to-one mapping of a pixel in a camera to a pixel in another camera, hence creating a relation between multiple cameras to enable reasoning between themselves. \\
Surround-view images can be represented by ${\mathbf{I} \in \mathbb{R}\textsuperscript{N×V×H×W×3}}$. Here, N,V, H and W are the number of temporal-frames, number views, height and width respectively. 
\subsection{Why surround-view in an AV?}
\begin{figure}[ht]
\vskip 0.2in
\begin{center}
\centerline{\includegraphics[width=\columnwidth]{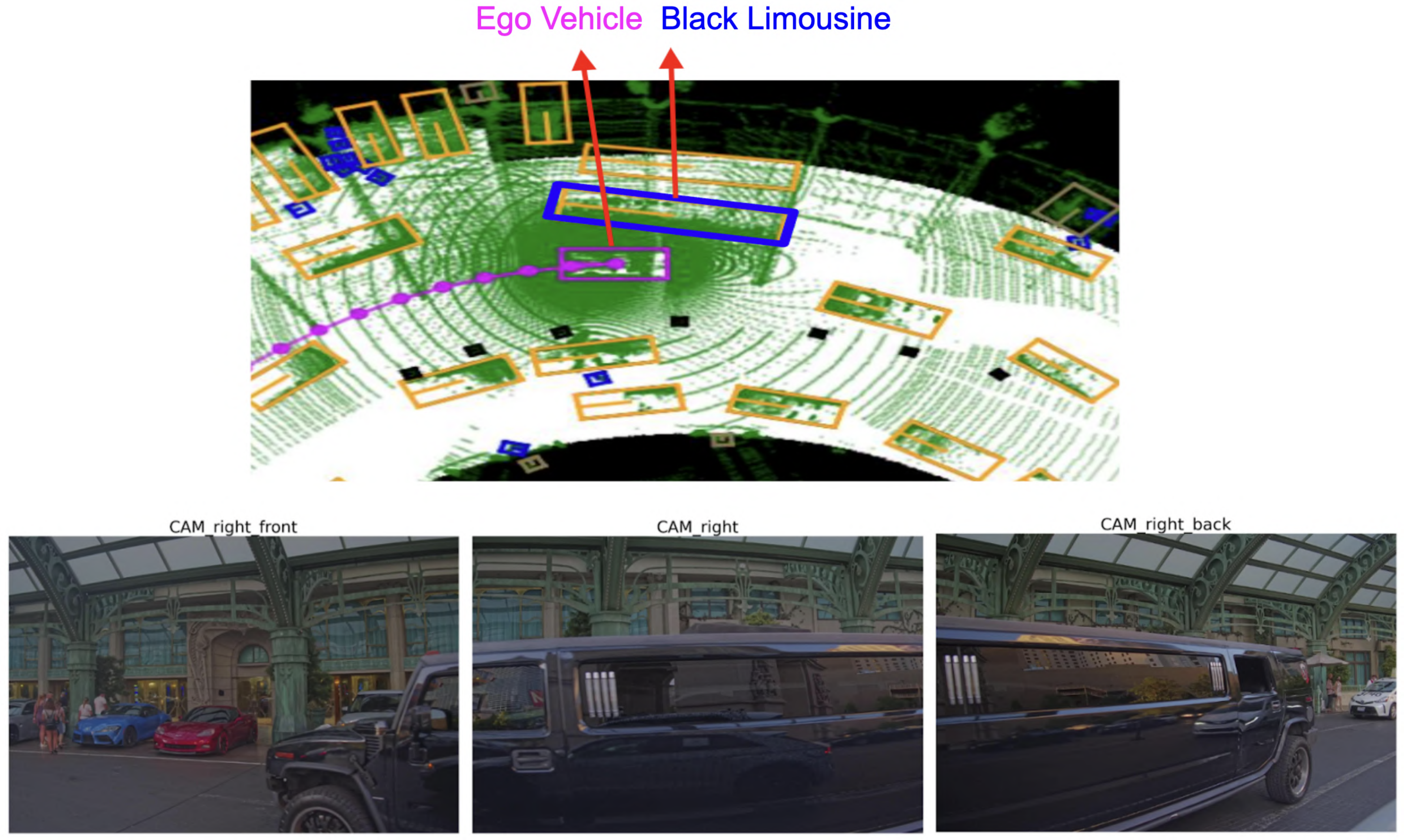}}
\caption{Usage of surround-view images in 3D object detection problem. BEV view (top); surround-view images of right-front, right, right-back cameras (bottom). This shows that with one or two of the cameras we may classify object as car, but without all the three images we won't be able to perfectly localize i.e. fit a bounding box on the black limousine.}
\label{limo}
\end{center}
\vskip -0.2in
\end{figure}

Lot of times it is hard to fit an entire object in the single frame to accurately detect and classify it. This is a specially common issue with the long vehicle category. Let’s take a visual understanding of what this means in as \cref{limo}

\subsection{Types of Surround-view Detectors}
SoTA surround-view Detection can be broadly classified amongst two subgroups viz., \emph{Geometry based view transformers} and \emph{Cross-attention based vision-transformers}.
\subsubsection{View Transformers}
\begin{figure}[ht]
\vskip 0.2in
\begin{center}
\centerline{\includegraphics[width=\columnwidth]{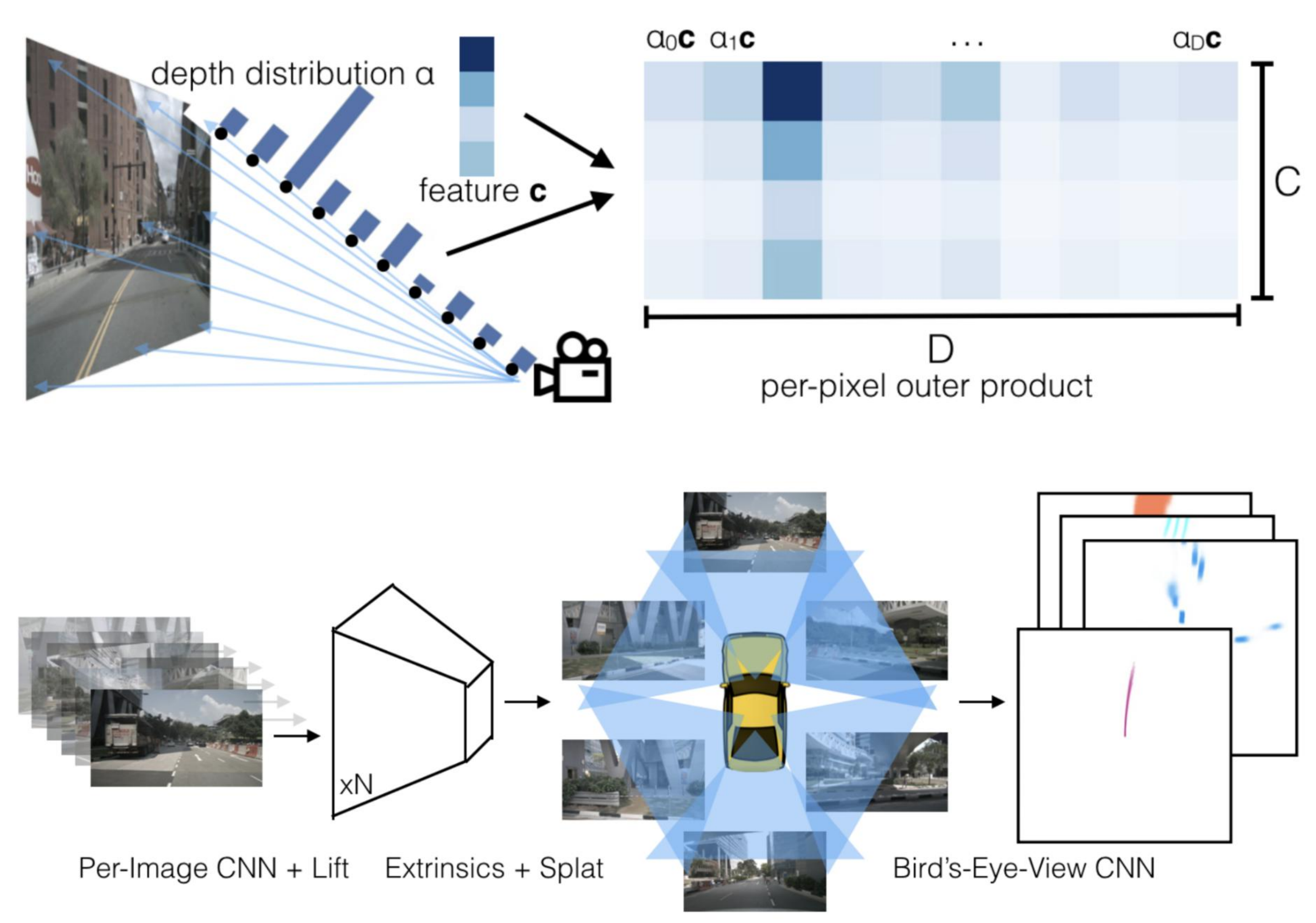}}
\caption{Lift-splat-shoot (LSS) \cite{lss_20} architecture: Lift step is visualized where per-image-frustum's pixel is projected to a discrete depth in BEV coordinate space with a context vector(top). Overall architecture is shown which takes in n images and returns BEV semantic map (bottom).}
\label{lss}
\end{center}
\vskip -0.2in
\end{figure}

Pioneer work: Lift, Splat, Shoot \cite{lss_20} started a chain work where they \emph{lift} each image individually into a frustum of BEV features, then \emph{splat} all frustums onto a rasterized BEV grid. Given \emph{n} images  ${\mathbf{X_{k}} \in \mathbb{R}\textsuperscript{3xHxW}}_n$, each with an extrinsics matrix ${\mathbf{E_k} \in \mathbb{R}\textsuperscript{3x4}}$ and an intrinsics matrix ${\mathbf{I_k} \in \mathbb{R}\textsuperscript{3x3}}$, we can find a rasterized BEV map of the feature in BEV coordinate frame as ${\mathbf{y} \in \mathbb{R}\textsuperscript{CxXxY}}$, where C, X, and Y are channel depth, height and width of the BEV map. The extrinsic and intrinsic matrices together define the mapping from reference coordinates $(x, y, z)$ to local pixel coordinates $(h, w, d)$ for each of the n cameras. This approach do not require access to any depth sensor during training or testing, just 3D box annotations are enough.  This architecture is demonstrated in as \cref{lss}. One of the latest development on this line of work is BEVDet \cite{bevdet_21}, which improves on pre-processing and post-processing techniques.

BEVDet4D \cite{bevdet4d_22} adds temporal dimensionality to this method and makes it a 4-dimensional problem. They tried to address inherent problem of high velocity error in vision-based detectors. Single-frame vision-based detectors generally have higher velocity errors compared to the laser-based sensors as LiDAR detectors generally uses multiple-sweep data with temporal information embedded in the point cloud; and RADAR's inherent point cloud includes velocity attribute using Doppler effect. Adding temporal frames in vision-detector enables us to learn temporal cues of the dynamic agent on the road. \\ \\
As a further extension, BEVDepth \cite{bevdepth_22} method adds camera-aware depth estimation module which facilitates the object depth predicting capability. Their hypothesis is that \emph{enhancing depth is the key to high-performance camera 3D detections} on nuScenes benchmark. They have replaced vanilla segmentation head in LSS with the CenterPoint \cite{centerpoint_20} head for 3D detection. For the auxiliary depth head baseline they use supervision from the detection loss only. However, due to the difficulty of monocular depth estimation, a sole detection loss is far from enough to supervise the depth module. Then used calibrated LiDAR data to project point cloud on to the images using camera transformation matrices hence forming a 2.5D image coordinates
$P^{img}i(u, v, d)$, where u and v denote coordinates in pixel coordinate and d denotes depth from the corresponding LiDAR point-cloud. To reduce memory usage, further development of $M^2BEV$ \cite{M2BEV_22} decreases the learnable parameters and achieves high efficiency on both inference speed and memory usage. \\ \\
These detectors include four components: 1. An image encoder to extract the image features in perspective view; 2. A depth module to generate depth and context, then outer product them to get point features; 3. A view transformer to convert the feature from perspective view to the BEV view; and lastly 4. A 3D detection head to propose the final 3D bounding boxes. BEVStereo \cite{bevstereo_22} introduces dynamic temporal stereo method to enhance depth prediction within compute cost-budget. Simple-BEV \cite{simple_bev_22} introduces RADAR point cloud on LSS approach. Based out of view transformers, BEVPoolv2 \cite{bevpool_v2_22} is the current SOTA as per nuScenes \cite{nuscenes_20} vision-detection leaderboard. They use BEVDet4D based backbone with dense-depth and temporal information for training. They have shown \emph{TensorRT} runtimes speedups as well. \emph{TensorRT} is the model format generally used by \emph{Nvidia} deployment hardware. 

\subsubsection{Vision Transformers}
\begin{figure}[ht]
\vskip 0.2in
\begin{center}
\centerline{\includegraphics[width=\columnwidth]{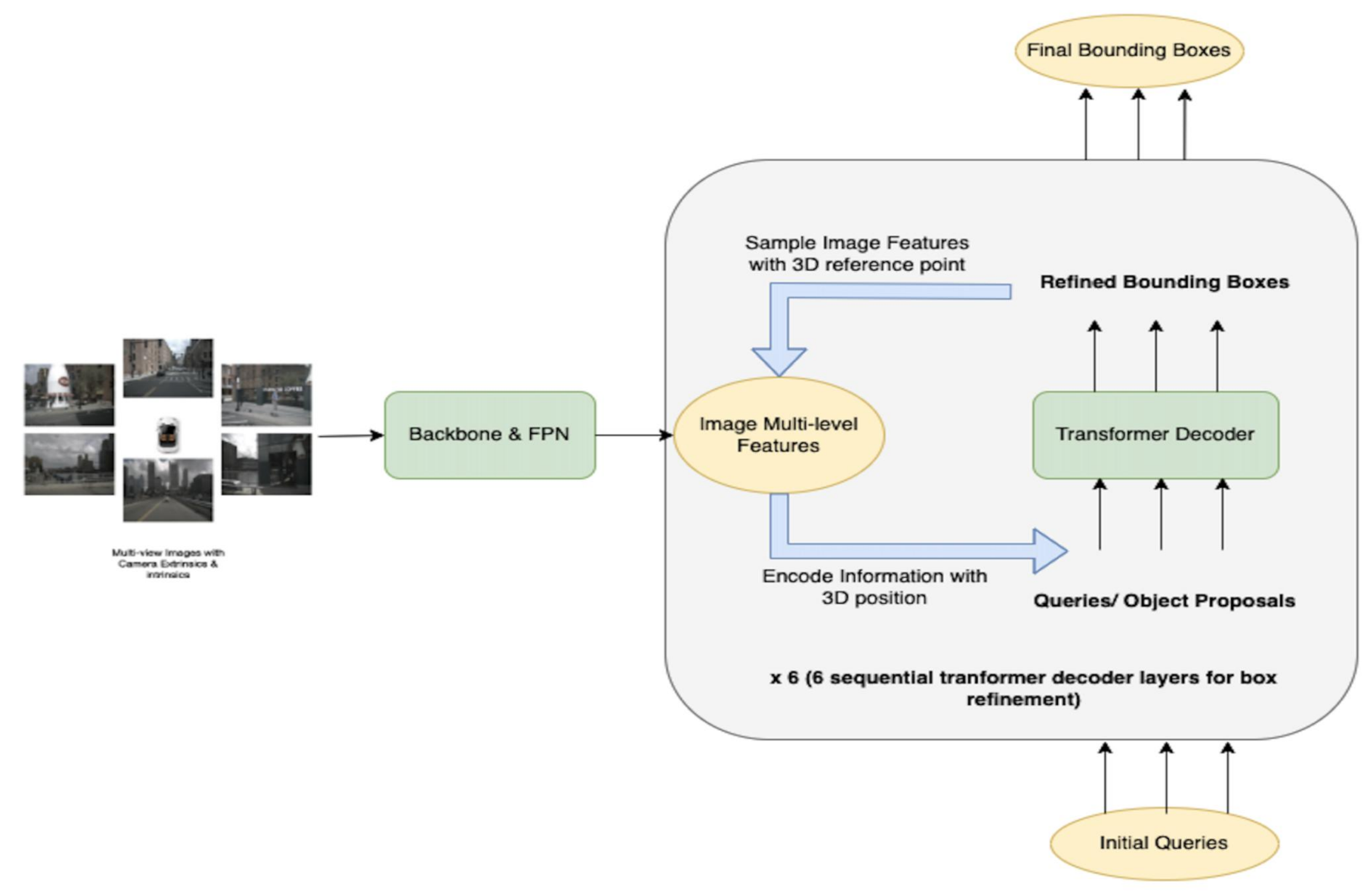}}
\caption{Adaptation workflow from DETR3D \cite{detr3d_21}.}
\label{svisn}
\end{center}
\vskip -0.2in
\end{figure}

Vision Transformers can be divided as per the granularity of the queries (object proposals) in the transformer decoder as per \cite{bev_survey_22} viz., sparse query-based and dense query-based methods. We will go into details about the representative work for both of these categories: \\ \\
\textbf{Sparse Query-based ViT:} In this line of work, we try to learn object proposals to look for in the scene, from the representative training data, and then use those learned object proposals to query at the test-time. Here assumption is made that test data objects are representative of the training data ones.  \\ \\
Seminal paper with single-image (Perspective-view), DETR \cite{detr_20} started this line of work, which is later extended to surround-view images in BEV space with DETR3D \cite{detr3d_21}. Here given $n$ surround-view images ${\mathbf{I} \in \mathbb{R}\textsuperscript{H'×W'×3}}$, the backbone and/or FPN and/or Transformers encoder produce $n$ encoded image features ${\mathbf{F} \in \mathbb{R}\textsuperscript{HWxd}}$, where d is the feature dimension, and H', W' and H,W denote spatial sizes of the image and the features, respectively. Then these $n$ encoded features and a small set of object queries ${\mathbf{Q} \in \mathbb{R}\textsuperscript{Nxd}}$ are fed into the Transformer decoder to produce detection results. Here $N$ is the number of the object queries, typically $300\sim900$ for the entire $360^\circ$ scene. As a meta-data camera transformation matrices is also used as an input. These matrices are required to create 3D reference point mapping onto the 2D coordinate-space and sample respective 2D-features per-query. \\ \\
In the Transformers decoder, object queries are sequentially processed by a self-attention module, a cross-attention module, and a feed-forward network (FFN), and then finally by a Multi-Layer Perceptron (MLP) to produce 3D BEV detections as the final output. For an interpretation: object queries denote potential objects at different locations on the BEV map; the self-attention module performs message passing among different object queries; and in the cross-attention module, object queries first search for the corresponding regions/ views to match, then distill relevant features from the matched regions for the subsequent predictions. Also worth noting, transformer-based encoder is an optional add-on here, but the core part of these detectors is in the transformer-based decoders. Workflow of this approach can be easily understood in as \cref{svisn}

As a further development of this work, Polar DETR \cite{polar_detr_21} parameterizes 3D detections in polar coordinates, which reformulates position parametrization, velocity decomposition, perception range, label assignment and loss function in polar coordinate system $(r, \theta)$. This approach eases optimization and enables center-context feature aggregation to enhance the feature interaction. In Graph-DETR3D \cite{graph_detr3d_22} they quantify the objects located at different regions and find that the “truncated instances” (i.e.,at the border regions of each image) are the main bottleneck hindering the performance of DETR3D. Although it merges multiple features from two adjacent views in the overlapping regions, DETR3D\cite{detr3d_21} still suffers from insufficient feature aggregation, thus missing the chance to fully boost the detection performance. To address this issue Graph-DETR3D\cite{graph_detr3d_22} aggregates surround-view imagery information through graph structure learning (GSL). It constructs a dynamic 3D graph between each object query and 2D feature maps to enhance the object representations, especially at the image-border regions. \\ \\
In a positional encoding development work by PETR \cite{petr_22} cites problem with 2D encoding of features in the former approach. They transform surround-view features into 3D domain by encoding the 3D coordinates from camera transformation matrices. Now, object queries can be updated by interacting with the 3D position-aware features and generate 3D predictions, hence making the procedure simpler. A follow-up work PETRv2 \cite{petr_v2_22} adds temporal dimensionality to it to get temporal-aware denser features.

\begin{table*}[t]
\caption{Results of vision-only 3D object detections on nuScenes camera-only 3D detection benchmark on test set. Abbreviations are defined in \cref{experiments_key}. }
\label{experiments}
\vskip 0.15in
\begin{center}
\begin{small}
\begin{sc}
\begin{tabular}{lcccccccccr}
\toprule
Method & Year & mAP & mATE & mASE & mAOE & mAVE & mAAE & NDS\\
\midrule
BEVPoolv2 & 2022 & 0.586 & 0.375 & 0.243 & 0.377 & 0.174 & 0.123 & 0.664 \\
BEVFormer v2 & 2022 & 0.580 & 0.448 & 0.262 & 0.342 & 0.238 & 0.128 & 0.648 \\
BEVStereo & 2022 & 0.525 & 0.431 & 0.246 & 0.358 & 0.357 & 0.138 & 0.610 \\ 
BEVDepth & 2022 & 0.503 & 0.445 & 0.245 & 0.378 & 0.320 & 0.126 & 0.600 \\ 
PolarFormer & 2022 & 0.493 & 0.556 & 0.256 & 0.364 & 0.439 & 0.127 & 0.572 \\
PETR v2 & 2022 & 0.490 & 0.561 & 0.243 & 0.361 & 0.343 & 0.120 & 0.582 \\ 
BEVFormer & 2022 & 0.481 & 0.582 & 0.256 & 0.375 & 0.378 & 0.126 & 0.569 \\
BEVDet4D & 2022 & 0.451 & 0.511 & 0.241 & 0.386 & 0.301 & 0.121 & 0.569 \\ 
Graph-DETR3D & 2022 & 0.425 & 0.621 & 0.251 & 0.386 & 0.790 & 0.128 & 0.495 \\
PolarDETR & 2022 & 0.431 & 0.588 & 0.253 & 0.408 & 0.845 & 0.129 & 0.493 \\
BEVDet & 2021 & 0.424 & 0.524 & 0.242 & 0.373 & 0.950 & 0.148 & 0.488 \\
PETR & 2022 & 0.434 & 0.641 & 0.248 & 0.437 & 0.894 & 0.143 & 0.481 \\
DETR3D & 2021 & 0.412 & 0.641 & 0.255 & 0.394 & 0.845 & 0.133 & 0.479 \\
FCOS3D & 2021 & 0.358 & 0.690 & 0.249 & 0.452 & 1.434 & 0.124 & 0.428 \\
CenterNet & 2019 & 0.338 & 0.658 & 0.255 & 0.629 & 1.629 & 0.142 & 0.400 \\
\bottomrule
\end{tabular}
\end{sc}
\end{small}
\end{center}
\vskip -0.1in
\end{table*}

\begin{figure}[ht]
\vskip 0.2in
\begin{center}
\centerline{\includegraphics[width=\columnwidth]{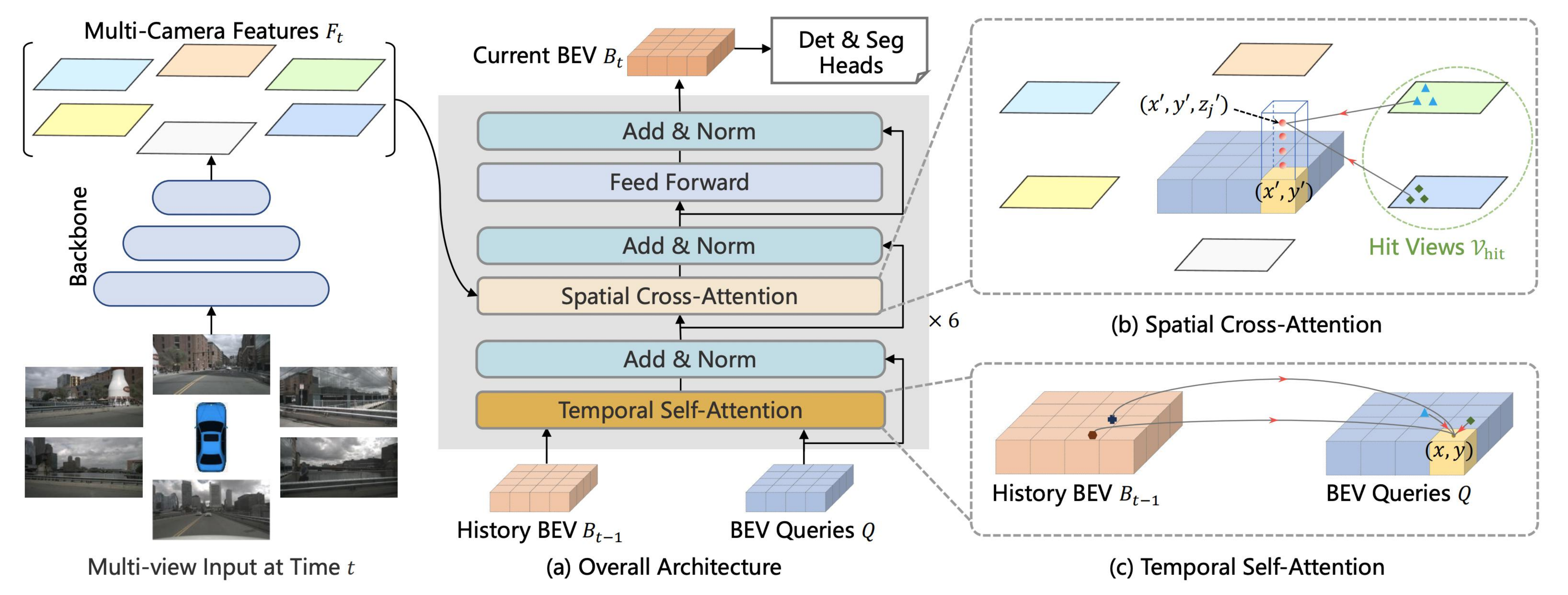}}
\caption{Overall architecture of BEVFormer \cite{bevformer_22}. (a) The encoder layer of BEVFormer contains
grid-shaped BEV queries, temporal self-attention, and spatial cross-attention. (b) In spatial crossattention, each BEV query only interacts with image features in the regions of interest. (c) In temporal self-attention, each BEV query interacts with two features: the BEV queries at the current timestamp and the BEV features at the previous timestamp.}
\label{bevformer}
\end{center}
\vskip -0.2in
\end{figure}

\textbf{Dense Query-based ViT:}
Here we have a dense-query based on region of interest in the BEV representation. Each query is pre-allocated with a spatial location in 3D space. This line of work is better than former in the sense that we will still be able to detect certain type of objects that were not learnt as object proposals in the training data with sparse-query. In other words, \emph{this approach is more robust to the scenario when training data is not the perfect representative of the test data}. \\ \\
Pioneer work with this line of work was BEVFormer \cite{bevformer_22}. They exploit both spatial and temporal information by interacting with spatial and temporal space through predefined grid-shaped BEV queries. To aggregate spatial information, they designed spatial cross-attention that each BEV query extracts from spatial features across the camera views. For temporal information, they use temporal self-attention to recurrently fuse the history BEV information as shown in \cref{bevformer}. This approach at the time have surpassed sparse-query based Vision Transformers methods by getting higher recall values, owing to the fact of exploiting dense queries. However, dense queries come at the cost of high compute requirement, which was tried to address using deformable-DETR's \cite{deformable_20} K-points around reference point sampling strategy. The fully transformer-based structure of BEVFormer makes its BEV features more versatile than other methods, easily supporting non-uniform and non-regular sampling grids.

A follow-up work, BEVFormerV2 \cite{bevformer_v2_22} adds perspective supervision which helps in convergence and leverages image-based backbone in a better manner. This brings back two-stage detectors, where proposals from the perspective head are fed into the bird’s-eye-view head for the final predictions. In addition to the perspective head proposals they also use DETR3D style learned queries. For auxiliary perspective loss, they use FCOS3D \cite{fcos3d_21} head which predicts the center location, size, orientation, and projected center-ness of the 3D bounding boxes. The auxiliary detection loss of this head, denoted as perspective loss $L_{pers}$, serves as the complement to the BEV loss $L_{bev}$, facilitating the optimization of the backbone. The whole model is trained with a total objective
\begin{equation}
L_{total} = \lambda_{bev}L_{bev} + \lambda_{pers}L_{pers}
\end{equation}

PolarFormer \cite{polarformer_22} reasons that nature of the ego car’s perspective, as each onboard camera perceives the world in shape of wedge intrinsic to the imaging geometry with radical (non-perpendicular) axis. Hence they advocate the exploitation of the Polar coordinate system on top of the BEVFormer \cite{bevformer_22}. 

\section{Experiments}
\label{experiments_key}
nuScenes \cite{nuscenes_20} is the widely used datasets in the literature for which sensor setup shown in \cref{sensor_setup} includes 6 calibrated cameras covering the entire $360^{\circ}$ scene. Results on discussed pioneer works are shown on the test set of nuScenes in \cref{experiments}. This is under the filter \emph{camera track detections}. The key for the metric abbreviations is as follows: mAP: mean Average Precision; mATE: mean Average Translation Error; mASE: mean Average Scale Error; mAOE: mean Average Orientation Error; mAVE: mean Average Velocity Error; mAAE: mean Average Attribute Error; NDS: nuScenes detection score.

\section{Further Extensions}
\label{further_extension_section}
Based on the most-recent developments around the surround-view BEV vision detections, we will now highlight possible future directions for the research. \\ \\
\textbf{Deployment compute-budget and run-time constraints:} Autonomous vehicle operate on a tight compute budget, as there is a limit of compute resources we can have on-board. However, when 5G internet becomes mainstream and all computation could have been shifted to the cloud computers. We, industry as a whole should start focusing on run-time constraints of these compute-expensive transformer based networks. One possible direction is to limit the object-proposals (queries) based on input-scene constraints. However, there is a need to have a smart way to handle it, else these networks may suffer through a low recall issue.\\ \\
\textbf{Smart object proposal initialization strategies:} We may come with a query initialization strategies that mix-and-match sparse and dense query initialization to enable pros of both. Major con of dense query-based approach is their high run-time. This can be handled by the usage of HD-maps to focus only on the areas of the road which matter the most. Just like BEVFormerv2 \cite{bevformer_v2_22}, object proposals can also be take from different modalities. As a one step further, these proposals may also be taken from the past time-step, with a fair assumption that driving scene won't have changed much within a fraction of a second. However, in order to make AVs scalable, researchers need to focus more on affordable sensors like cameras and RADARs and not too much on expensive LiDARs or HD-Maps. \\ \\ 
\textbf{Collaborative Perception:} A relatively new field of area is how to make use of multi-agents, mutli-view transformers to enable collaborative perception. This setup requires a minimal infrastructure setup to enable smooth communications between different AVs on the road. CoBEVT \cite{cobevt_22} shows initial proof of how Vehicle-to-Vehicle communication may lead to superior perception performance. They test their performance on OPV2V \cite{OPV2V_21} benchmark dataset for V2V perception.

\section{Conclusion}
\label{conclusion_section}
In this work we introduced development work around vision-based 3D object detection focused on autonomous vehicles. We went through more than $60$ papers and $5$ benchmark datasets to prepare this paper.\\
To be specific, we first build a case on why camera based surround-view detection head is important for solving autonomous vehicles. Then we started off with how research has progressed from single-view detection and extended to surround-view detection head paradigm and thereby increasing the detection performance. We have categorized two prominent categories for surround-view camera detectors to keep an eye on viz., {Geometric View-Transformers based} and \emph{Vision Transformers based}. In the end we proposed our take on surround-view detection trends with the focus of deploying those networks on an actual autonomous car, which may enlighten future research work.

\bibliography{example_paper}
\bibliographystyle{icml2023}



\end{document}